\documentclass{article}
\usepackage[round]{natbib}
\usepackage[preprint]{neurips_2024}
\usepackage[utf8]{inputenc} 
\usepackage[T1]{fontenc}    
\usepackage{hyperref}       
\usepackage{url}            
\usepackage{booktabs}       
\usepackage{amsfonts}       
\usepackage{nicefrac}       
\usepackage{microtype}      

\usepackage{lipsum}
\usepackage{fancyhdr}       
\usepackage{graphicx}       
\graphicspath{{media/}}     
\usepackage{amsmath}
\usepackage{float}
\usepackage{algorithm}
\usepackage{algpseudocode}
\usepackage{graphicx}
\usepackage{subcaption} 
\usepackage[table]{xcolor}
\usepackage{multirow}
\usepackage[most]{tcolorbox} 

\title{Chained Recursive Language Models for Multi-Iteration Reasoning}

\author{
  Purbesh Mitra \\
  University of Maryland \\
  \texttt{pmitra@umd.edu} \\
   \And
  Sennur Ulukus \\
  University of Maryland \\
  \texttt{ulukus@umd.edu} \\
}

\begin{document}
\maketitle

\begin{abstract}
Long context reasoning in large language models (LLMs) is usually constrained by the fact that a single inference trajectory has to simultaneously explore the context, store intermediate state, verify evidence, and produce the final answer. This becomes particularly difficult in tasks that require extraction, counting, ordering, or multi-hop reasoning, where an early mistake can propagate until the final response. In this work, we propose \textbf{Chained Recursive Language Models (Chained RLM)}, an inference-time architecture, in which the same underlying model is called repeatedly as a sequence of fresh reasoning roots. Each root receives the original problem and context, but does not inherit the full conversational history. Instead, it receives a compact plain-text summary, a plain-text blackboard, and some durable task-specific artifacts written by predecessor roots. The motivation is to manage the context by chopping into partial tasks rather than one large inference response; in each  staged computation, intermediate artifacts can be inspected, corrected, and extended by a later fresh inference by the same model. We describe the system model, handoff mechanism, artifact workspace, and evaluation protocol for this system. We study when fresh-context artifact continuation gives a measurable gain in accuracy over direct LLM answering even with recursive tool-calling.
\end{abstract}

\section{Introduction}
Recently, long context reasoning in large language models (LLMs) have become a central direction in building agentic systems which can work over large documents, codebases, transcripts, and multi-step data~\citep{plaat2025agentic}. The most straightforward idea is to put all the necessary information, including the prompt in the context and ask the LLM to reason over it directly. This simple approach is powerful, but it has some important limitation: the entire problem solving process has to occur inside one interaction trace. The model has to read the context, identify relevant evidence, keep track of unresolved assumptions, perform any computation or tool use, and decide when the answer is ready. In short contexts, this is often sufficient. In long contexts, however, the model can easily lose the structure of the work. It may perform a shallow extraction, forget what was already checked, or finalize an answer before a real audit has happened. This is also known as \emph{context rot}~\citep{hong2025context}.

There are several existing ideas for improving inference-time reasoning. Chain-of-thought prompting elicits intermediate reasoning traces~\citep{wei2022chain}, self-consistency samples multiple reasoning paths~\citep{wang2022self}, and Tree-of-Thoughts explores alternatives through deliberate search~\citep{yao2023tree}. Tool-using language agents, such as ReAct, interleave reasoning with actions to gather evidence from an external environment~\citep{yao2022react}. Memory-based systems such as MemGPT manage information beyond the active context window by moving information between memory tiers~\citep{packer2023memgpt}. Reflexion stores verbal feedback from previous attempts as an episodic memory buffer~\citep{shinn2023reflexion}. MOTIF~\citep{mitra2025motif} decomposes task into multiple modules as solve them iteratively. Recursive language models~\citep{zhang2025recursive} call different parts of the context for recursive model calls with tool-use in a total context path. These approaches share a common drawback: a single forward pass is often not the right method for difficult reasoning tasks.

In this work, we explore a different but related direction. Instead of asking one model call to solve the entire problem, we call the same recursive language model multiple times in a linear chain: \textbf{chained RLM}. We call each RLM call a \emph{root}. A root is a fresh instance of the same model and recursive context based tool-calling environment. It receives the original problem and context, but only a compact continuity state from previous roots. This continuity state is intentionally simple: a plain-text summary, a plain-text blackboard, and a set of plain-text artifacts saved on disk. Raw predecessor trajectories are also saved, but they are not automatically injected into every prompt. A later root may inspect them only if the current summary or artifact is insufficient.

The key intuition behind this artifact-based architecture is that long context reasoning is a burden for state management over the context-size. A model may have enough context length to see the whole document, but still fail to maintain a clean working state to steer its reasoning towards the correct flow. Our hypothesis is that durable artifacts can provide a more stable interface between reasoning stages. For example, in a counting task, Root 0 can construct a candidate event ledger. Root 1 can inspect and correct this ledger, rather than repeating the entire extraction from scratch. Root 2 can then audit the final counts and answer. If this works, the chain acts as an inference-time analog of a staged data analysis pipeline, where intermediate outputs are visible and editable.

Our system is not trained. It does not use reinforcement learning, and it does not use a specialized verifier by default. It relies on the same LLM being called in fresh roots, with continuity passed through human-readable text. However, using different RL environments on top of this architecture the model can be more familiarized with the Chained RLM structure.

Our contributions are summarized as follows:

\begin{itemize}
    \item We propose Chained RLM, a linear inference-time architecture where the same model solves a long-context task through multiple fresh roots.
    \item We introduce a simplified handoff mechanism based on plain-text summary, blackboard, and next-action fields, avoiding JSON or host-parsed state.
    \item We introduce a plain-text artifact workspace where roots create, read, and maintain task-specific evidence ledgers, derivations, audits, and checklists.
    \item We define a system model and evaluation protocol for comparing Chained RLM against regular LLM baselines under controlled model and task settings.
\end{itemize}

\begin{figure}
    \centering
    \includegraphics[width=\linewidth]{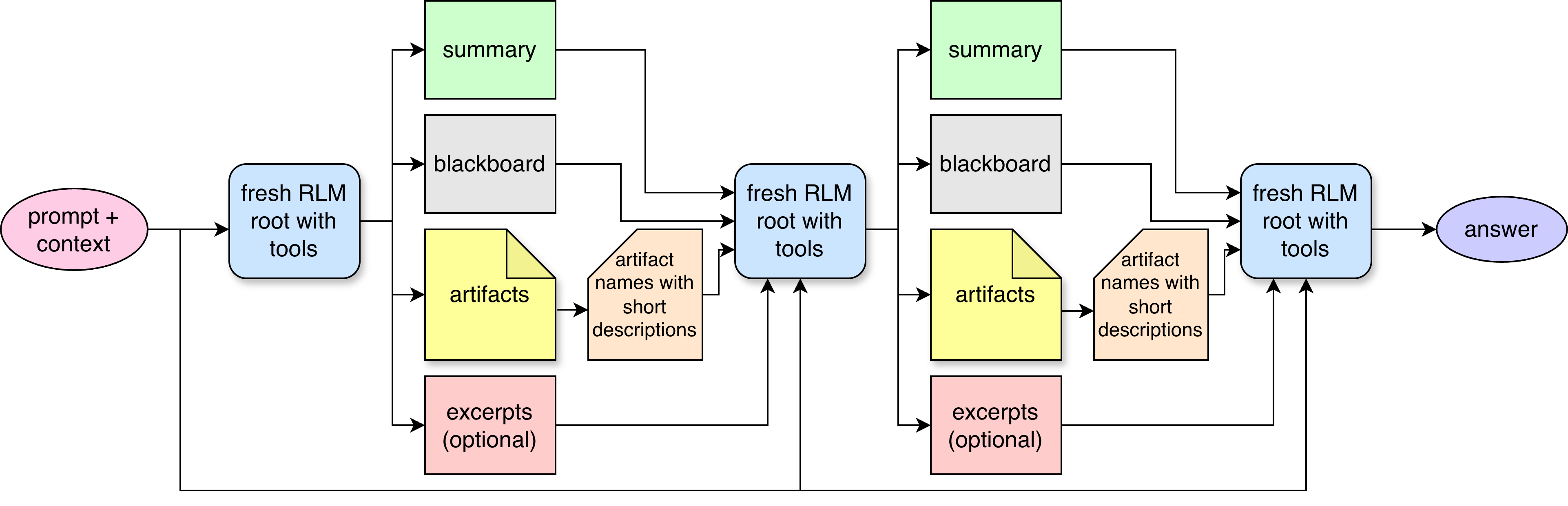}
    \caption{Chained RLM architecture with 3 RLM calls.}
    \label{fig: Chained_RLM}
\end{figure}

\section{Background and Related Works}
LLM reasoning has often been improved by adding more inference-time structure. Chain-of-thought prompting~\citep{wei2022chain} shows that intermediate reasoning traces can improve performance on reasoning problems. Self-consistency~\citep{wang2022self} further improves this by sampling multiple solution paths and selecting the most consistent answer. Tree-of-Thoughts~\citep{yao2023tree} generalizes this idea by making reasoning states into units that can be searched over. These methods mostly treat reasoning as a search or sampling problem over text.

Another line of work considers language models as agents that can interact with external tools or environments. ReAct~\citep{yao2022react} interleaves reasoning traces and actions, allowing the model to query external information sources or perform environment steps. DSPy~\citep{khattab2023dspy} provides a programming model for composing language model calls into structured pipelines, and is closely related to our implementation perspective. In our work, each root is a tool-using RLM-style module that can inspect variables, run code, call sub-LLMs, and write artifacts.

Memory is another important component for long-context tasks. Retrieval-augmented generation~\citep{lewis2020retrieval} combines parametric model knowledge with external non-parametric memory. MemGPT~\citep{packer2023memgpt} frames long-context interaction as a memory management problem, where information is moved between active context and longer-term storage. Reflexion~\citep{shinn2023reflexion} stores verbal feedback from previous trials, and then uses this memory to improve later attempts. MOTIF~\citep{mitra2025motif} decomposes a prompt into multiple task modules and solves them iteratively, where the accuracy of a taskis considered at the final answer formulation.

Recursive Language Models (RLMs) treat long prompts as an external environment and let a root model inspect, decompose, and recursively call itself over relevant input parts~\citep{zhang2025recursive}. This improves long-context inference without changing model weights, but the global working state still largely lives inside one orchestration trajectory. Early extraction or aggregation mistakes can therefore become stale assumptions that later steps continue to use. Chained RLM addresses this limitation by passing a plain-text summary, blackboard, and durable artifacts to a fresh root of the same model, making intermediate state easier to inspect, edit, and audit.

Chained RLM can be described as Chained memory structure along with the RLM architecture. It differs from RLM in one important way: the memory is not just a retrieval storage. It is a set of task-specific artifacts that are generated dynamically by the model itself and the later roots are instructed to preserve, edit, and audit them. The motivation is that a single LLM trajectory can become internally stale. Once a model writes an intermediate conclusion, it may continue relying on it, even when the conclusion was produced by a weak extraction or a wrong event definition. A fresh root provides a natural opportunity to re-read the artifact as an external object. In this sense, Chained RLM is a method for creating checkpoint boundaries where the current working state can be inspected by the same model under a new context.

\section{System Model}

We consider a long-context task represented by a pair $(q, c)$, where $q$ is the user question and $c$ is the original context. The context can be a document, transcript, table, codebase, or synthetic long-context instance. A regular LLM baseline directly produces an answer
\begin{align}
    y \sim f_\theta(q, c),
\end{align}
where $f_\theta$ is the LLM with parameters $\theta$. In practice, the prompt may contain task instructions and output-format constraints. But all reasoning occurs inside one LLM call.

In Chained RLM, as shown in Fig.~\ref{fig: Chained_RLM}, we instead define a sequence of roots $r = 0, 1, \ldots, R-1$, where $R$ is the maximum chain length. Each root uses the same underlying model $f_\theta$ and receives the same original problem and context $(q,c)$. The only changing input is a chain state $s_r$:
\begin{align}
    o_r \sim f_\theta(q, c, s_r; \mathcal{T}),
\end{align}
where $\mathcal{T}$ denotes the available tools, and $o_r$ is either a final answer or a handoff to the next root.

The chain state $s_r$ is composed of:
\begin{align}
    s_r = \left(B_r, A_r, H_r, E_r\right),
\end{align}
where $B_r$ is the current plain-text blackboard, $A_r$ is the set of artifacts available to the root, $H_r$ is a compact predecessor handoff summary, and $E_r$ is an optional recent-work excerpt from the previous raw trajectory. The original raw trajectory of each root is saved separately as $T_r$, but is not placed into the prompt by default. A later root can inspect $T_i$ for $i<r$ only if the summary or artifact is insufficient, contradictory, or unsupported.

\subsection{Roots}
A root is a fresh RLM call with access to a Python REPL-like environment. The root can inspect the original context, run code, print intermediate outputs, write artifacts, read artifacts, and submit either a final answer or a handoff. The important design choice is that a later root does not inherit the conversational history of the previous root. It has the same original inputs, but only the curated continuity state. This is meant to reduce the effect of stale reasoning while still preserving useful work.

\subsection{Blackboard}
The blackboard is a plain-text working memory. It contains the current best answer, verified facts, assumptions, open questions, contradictions, evidence pointers, and reusable procedures. The host does not parse the blackboard into structured fields. This is a deliberate simplification. The blackboard is written for the next fresh root, not for the host program.

\subsection{Artifacts}
Artifacts are persistent plain-text files created by roots. They can be candidate ledgers, extraction tables, derivations, audit notes, or verification checklists. Unlike the blackboard, artifacts can be detailed. The types and numbers of artifacts are generated by the root RLM itself. For example, a counting task may use an artifact with one row per candidate event. A later root should preserve the structure of the artifact while correcting, extending, or auditing it. In this way, the artifact becomes the main object of continuation.

\subsection{Handoff}
If a root does not produce a final answer, it submits a handoff. The handoff is in plain text with exactly three sections. However, an optional fourth section of recent work excerpts can be added.
\begin{tcolorbox}[colback=gray!10!white,
                  colframe=gray!80!black,
                  arc=2mm,
                  boxrule=0.5pt,
                  left=1mm, right=1mm, top=1mm, bottom=1mm]
\textbf{Handoff Format:}
If you do not submit FINAL, submit exactly this plain-text format:

\begin{verbatim}
HANDOFF:
SUMMARY:
What this root did, what it found, and what is currently believed.

BLACKBOARD:
The current compact working state for the next root.

NEXT:
One concrete next action for the next root. It must name what to inspect or
compute and why it can change, verify, or falsify the answer.
\end{verbatim}

\textbf{Recent Work Excerpt.}
If present, treat it as raw scratch flow only. It is not authoritative. Verify
important claims against the blackboard, artifacts, original context, or full
raw trajectory before relying on them.

\textbf{Final Format.}
If the answer is fully supported, submit:

\begin{verbatim}
FINAL: <answer>
\end{verbatim}

If this is the last root, submit the best supported FINAL answer.
\end{tcolorbox}
We do not use JSON, XML, YAML, or a host-parsed schema for the handoff. This avoids making the architecture depend on fragile formatting constraints, and keeps the continuity state closer to how a researcher would write notes to their future self.

\subsection{Overall System Prompt}
The overall system prompt for any root LLM call is the following: 
\begin{tcolorbox}[
  colback=gray!5!white,
  colframe=gray!70!black,
  title={Chained RLM Root Instruction},
  fonttitle=\bfseries,
  boxrule=0.5pt,
  arc=1mm,
  left=1mm,
  right=1mm,
  top=1mm,
  bottom=1mm
]
\small
You are a fresh root for the same task and the same underlying LLM/model.

The original problem and context are available to every root. Prior interaction
history is absent except for the plain-text blackboard, artifacts, predecessor
summaries, and optional recent-work excerpt.

Use the Python REPL iteratively. Variables \texttt{problem}, \texttt{context},
and \texttt{chain\_info} are already available. First inspect
\texttt{chain\_info} or call \texttt{chain\_status()}; task-specific rules live
there. Use short code steps. Call \texttt{SUBMIT} only when ready.

\textbf{Chain Rules}
\begin{enumerate}
    \item Use blackboard and artifacts first.
    \item Do one bounded useful step with tools/code.
    \item You MUST write or update one plain-text artifact before FINAL or HANDOFF.
    \item Root 0 chooses artifact structure; later roots preserve and update it.
    \item For long-context extraction, counting, or ordering tasks, Root 0 usually HANDOFFs after creating candidate rows. FINAL only after an audited artifact exists.
    \item Before FINAL, audit any relevant entity mapping, event definition, ordering, and count logic in the artifact.
    \item FINAL only when supported or when this is the last root.
    \item Otherwise HANDOFF as plain text using SUMMARY, BLACKBOARD, and NEXT.
\end{enumerate}

\textbf{Blackboard.}
The blackboard is plain text. It is not parsed by the host. Keep it short,
auditable, and useful to a fresh copy of the same LLM in the next node.

Recommended blackboard contents:
\begin{itemize}
    \item Current best answer
    \item Verified facts
    \item Assumptions / hypotheses
    \item Open questions
    \item Contradictions
    \item Evidence pointers
    \item Procedures / reusable checks
    \item Artifact names, purposes, structures, trust status, and changes made
\end{itemize}

Do not encode the blackboard as JSON. If a fact matters, write it clearly and
cite where it came from. If evidence is missing, say that explicitly.

\textbf{Artifacts.}
Artifacts are durable shared working memory across roots. They are not scratch
files. They let the same LLM/model in a later fresh node inspect and improve the
evidence without reading the whole raw trajectory.

Artifacts are for task-specific evidence too detailed for the blackboard:
indexes, candidate ledgers, tables, audit notes, derivations, extraction
procedures, contradiction logs, or verification checklists.

The model chooses artifact names and formats based on the question. Prefer
plain \texttt{.txt} or \texttt{.tsv}. Do not use JSON, XML, YAML, markdown
tables, or code fences.

Every root MUST create or update at least one question-specific plain-text
artifact before submitting HANDOFF or FINAL.

Root 0 MUST choose the initial artifact set and structure.

Every later root MUST first call \texttt{list\_artifacts()}, read the relevant
artifacts, preserve the existing structure, update/correct/audit/extend the
information inside that structure, and report exactly what changed in
BLACKBOARD.

Maintain means changing or correcting artifact information without sacrificing
the artifact's structure. If a structure must change, write a clearly named
superseding artifact, explain why the old structure failed, and record the
migration in BLACKBOARD.
\end{tcolorbox}
\section{Methodology}

\subsection{Root Execution}
Each root begins by inspecting the chain state. It reads the blackboard and artifact inventory, and then decides whether the current artifact state is sufficient for answering. If the task is a long-context extraction, counting, or ordering task, the first root is encouraged to create candidate rows and hand off, unless it already has an audited artifact. Later roots are encouraged to read existing artifacts before starting a new extraction. This instruction is important because otherwise the model may simply redo shallow analysis, which defeats the purpose of chaining.

\subsection{Artifact-Mediated Handoff}
The central mechanism this architecture is artifact-mediated handoff of teh resoning chain. Suppose the task asks for the most common event type in a long transcript. A direct LLM may count surface strings or hallucinate a category. In Chained RLM, a typical desired behavior is the following:
\begin{enumerate}
    \item Root 0 extracts candidate events and writes an artifact and list the artifact names.
    \item Root 1 reads the artifact, audits entity mappings and event definitions, and corrects the rows.
    \item Root 2 recomputes counts from the audited artifact and submits the final answer.
\end{enumerate}
This does not require the host to understand the task. The host only stores text. The LLM itself chooses the artifact format and maintains it across roots. It makes the system act more general in long-horizon task solving, rather than a specialized solver. 

\subsection{Finalization}
Before submitting a final answer, a root is instructed to audit any relevant entity mapping, event definition, ordering, and count logic in the artifact. This is a general instruction, not benchmark-specific. It applies to many tasks where the answer depends on extracting and aggregating evidence from long context. The last root is forced to submit the best supported final answer, but non-final roots should hand off when important uncertainty remains. The full algorithm is shown in Algo.~\ref{algo:chained_rlm}

\begin{algorithm*}
    \caption{Chained Recursive Language Model}
    \label{algo:chained_rlm}
    \begin{algorithmic}[1]
        \Require LLM $f_\theta$; problem $q$; context $c$; maximum roots $R$; tool set $\mathcal{T}$
        \State Initialize blackboard $B_0$ as plain text
        \State Initialize artifact set $A_0 \gets \emptyset$
        \State Initialize predecessor summaries $H_0 \gets \emptyset$
        \For{$r = 0,1,\ldots,R-1$}
            \State Build chain state $s_r = (B_r, A_r, H_r, E_r)$
            \State Run fresh root $o_r \sim f_\theta(q,c,s_r;\mathcal{T})$
            \State Save raw trajectory $T_r$ to disk
            \If{$o_r$ is \textsc{Final}}
                \State \Return final answer $y_r$
            \ElsIf{$o_r$ is \textsc{Handoff}}
                \State Parse plain-text \textsc{Summary}, \textsc{Blackboard}, and \textsc{Next}
                \State Update blackboard $B_{r+1}$ from handoff text
                \State Keep artifacts $A_{r+1}$ written by root $r$
                \State Append compact handoff summary to $H_{r+1}$
                \State Optionally append bounded recent-work excerpt $E_{r+1}$
            \EndIf
        \EndFor
        \State \Return best supported final answer from last root
    \end{algorithmic}
\end{algorithm*}

\section{Experiments}

\subsection{Settings}
We evaluate Chained RLM on long-context tasks where the answer cannot be reliably obtained by a single short pattern match. Candidate benchmark families include multi-hop retrieval, long-context aggregation, counting, and ordered event extraction. For each benchmark, we compare:
\begin{itemize}
    \item \textbf{Regular LLM}: direct prediction from the problem and context.
    \item \textbf{Chained RLM}: the proposed fresh-root artifact-mediated system.
\end{itemize}
Both systems use the same underlying model: GPT-5-mini. For Chained RLM, we record the maximum chain length, maximum iterations per root, maximum sub-LLM calls per root, and whether caching is enabled. For a fair comparison, the final experiment also includes a compute-matched setting, but in the first version we use the direct LLM baseline as a clean lower-complexity reference point.

\subsection{Evaluation Metrics}
The primary metric is the exact task accuracy. Since, LLMs are stochastically sampled, there are multiple definitions of accuracy measurements in benchmarks. For our work, we use the simplest pass@1 accuracy~\citep{chen2021evaluating}, which measures the probability of getting a correct answer in the first attempt. It is defined as 
\begin{align}
    \text{Pass@1 accuracy} = \frac{\text{\# correct answers}}{\text{\# questions}} = \frac{1}{L}\sum_{j=1}^{L}\mathbb{I}_j, 
\end{align}
where $L$ is the total number of questions in the benchmark; $\mathbb{I}_j = 1$, if the boxed answer to the $j$th question is in the response from the model, and $0$, otherwise.

We also measure the average number of roots used, handoff count, and cost of inference per task. These secondary metrics are crucial for determining the accuracy and resoyrce trade-off between the baseline LLM and the Chained RLM architecture.

\begin{table*}[t]
  \caption{Accuracy metrics reported for each selected benchmark.}
  \label{tab:main_results_balanced}
  \centering
  \begin{tabular}{ccc}
    \toprule
    BenchMark & Regular LLM (GPT-5-mini) toolcall & Chained RLM (with GPT-5-mini)\\
    \midrule
    RULER & 87\% & 92\% \\
    BABILong & 44\% & 59\% \\
    LongBench v2 & 41\% & 52\% \\
    OOLONG-real & 14\% & 38\% \\
    \bottomrule
  \end{tabular}
\end{table*}

\begin{table}[t]
\centering
\caption{Average resource use of regular LLM and Chained RLM. Token and cost
values are calculated based on the current API pricing of GPT-5-mini.}
\label{tab:resource_cost}
\small
\setlength{\tabcolsep}{4pt}
\begin{tabular}{llccccc}
\toprule
Benchmark
& Method
& Avg. \#
& Avg. \#
& Avg. \#
& Avg. \#
& Avg. Cost \\

& 
& Root Calls
& Handoffs
& Input Tokens
& Output Tokens
& per Task \\
\midrule
RULER
& Regular LLM
& 1.0 & 0.0 & 38k & 4k  & \$0.11 \\
& Chained RLM
& 1.8 & 0.8 & 68k & 11k & \$0.21 \\

\midrule
BABILong
& Regular LLM
& 1.0 & 0.0 & 52k & 6k  & \$0.13 \\
& Chained RLM
& 2.7 & 1.7 & 104k & 21k & \$0.32 \\

\midrule
LongBench v2
& Regular LLM
& 1.0 & 0.0 & 47k & 5k  & \$0.12 \\
& Chained RLM
& 2.4 & 1.4 & 91k & 20k & \$0.29 \\

\midrule
OOLONG-real
& Regular LLM
& 1.0 & 0.0 & 58k & 8k  & \$0.14 \\
& Chained RLM
& 3.4 & 2.4 & 132k & 35k & \$0.44 \\

\midrule
Average
& Regular LLM
& 1.0 & 0.0 & 48.8k & 5.8k  & \$0.125 \\
& Chained RLM
& 2.6 & 1.6 & 98.8k & 21.8k & \$0.315 \\
\bottomrule
\end{tabular}
\end{table}

\subsection{Comparison with Regular LLMs}

We compare Chained RLM against a regular LLM baseline on four long-context
benchmarks: RULER~\citealp{hsieh2024ruler},
BABILong~\citep{kuratov2024babilong}, LongBench v2~\citep{bai2025longbench},
and OOLONG-real~\citep{bertsch2025oolong}. Both methods use GPT-5-mini as the
underlying model and receive the same problem and context. The regular baseline
solves the task in a single inference trajectory, while Chained RLM is allowed
to use multiple fresh roots connected through plain-text handoffs, a blackboard,
and persistent task artifacts.

Table~\ref{tab:main_results_balanced} shows that Chained RLM improves accuracy
on all four benchmarks, with an average absolute gain of $13.75$ percentage points over the
regular LLM baseline. The Chained
system is more accurate, but the improvement is uneven across benchmarks.
The gain is relatively small on RULER, where many tasks can be solved by direct
retrieval or localized reasoning. In contrast, the improvement is much larger on
BABILong, LongBench v2, and especially OOLONG-real, where the model must preserve
partial evidence, compare events across distant parts of the context, or
aggregate many small observations before answering. This suggests that Chained
RLM is most useful when the main difficulty is not simply finding one relevant
span, but maintaining a reliable intermediate state over a long reasoning
process.

Table~\ref{tab:resource_cost} highlights the corresponding average resource use.
Chained RLM is more expensive because it uses more root calls, handoffs, and
tokens. The additional roots
give the system opportunities to externalize partial work into artifacts, audit
previous conclusions, and continue from a compact state instead of relying only
on the hidden state of one long generation. Therefore, the cost increase is due to the
the price of making intermediate reasoning more robust and interpretable.

The results therefore show a practical trade-off. For easier or mostly
retrieval-based tasks, a regular LLM may already be sufficient, and chaining may
not justify its additional cost. For tasks that require long-context record-keeping,
multi-step aggregation, or careful preservation of partial decisions, the
additional computation can produce a much larger accuracy gain.

\section{Discussion and Limitations}
The current architecture is intentionally simple, but this also creates limitations. First, the host does not enforce that a root must read artifacts before finalization. This keeps the method general, but can allow premature answers. Secondly, artifact quality is model-dependent. If Root 0 writes a flawed event definition, later roots may preserve that wrong structure instead of correcting it. Third, the whole chain can drift, i.e.,  a later root may ignore a good artifact and restart from a worse state. Another limitation is compute. A Chained system uses more model calls than a direct LLM baseline. Therefore, any gain must be interpreted together with the additional inference cost. 

\section{Conclusion}
In this work, we propose Chained Recursive Language Models (Chained RLM), an inference-time architecture for multi-iteration reasoning where the same model solves a task through a sequence of fresh root calls. The system passes on only plain-text continuity states: summary, blackboard, and artifacts. Raw predecessor trajectories are saved separately and inspected only when needed. This allows the durable artifacts making multi-iteration reasoning more like staged analysis, where intermediate work can be audited and improved by a later fresh model call. We define the system model, handoff mechanism, artifact workspace, and an initial evaluation protocol against regular LLMs. We show that the artifact-mediated fresh-context continuation is an improvement to the architecture of vanilla tool calling recursive LLMs.

\bibliography{references}

\end{document}